\def\year{2019}\relax
\documentclass[letterpaper]{article} 
\usepackage{aaai19}  
\usepackage{times}  
\usepackage{helvet}  
\usepackage{courier}  
\usepackage{url}  
\usepackage{graphicx}  
\usepackage{amsmath}
\usepackage{array}
\usepackage{multirow}

\def\ie{\textit{i.e.}}

\begin{document}
\title{Scale Invariant Fully Convolutional Network: Detecting Hands Efficiently}
\author{
	Dan Liu,\textsuperscript{\rm 1}
	Dawei Du,\textsuperscript{\rm 2}
	Libo Zhang,\textsuperscript{\rm 3}\thanks{Corresponding author: Libo Zhang (libo@iscas.ac.cn). This work was partially supported by the National Natural Science Foundation of China under Grant 61807033 and Grant 61771341, partially supported by US NSF IIS-1816227.}
	\\
	\Large \textbf { Tiejian Luo,\textsuperscript{\rm 1} 
		Yanjun Wu,\textsuperscript{\rm 3}
		Feiyue Huang,\textsuperscript{\rm 4}
		Siwei Lyu\textsuperscript{\rm 2} }
	\\ 
	\textsuperscript{\rm 1}University of the Chinese Academy of Sciences, China 
	\textsuperscript{\rm 2}University at Albany, SUNY, USA\\
	\textsuperscript{\rm 3}Institute of Software Chinese Academy of Sciences, China 
	\textsuperscript{\rm 4}Tencent Youtu Lab, China\\
	liudan171@mails.ucas.ac.cn, ddu@albany.edu, libo@iscas.ac.cn,\\ tjluo@ucas.ac.cn, yanjun@iscas.ac.cn, huangfeiyue@gmail.com, slyu@albany.edu
}  
\maketitle
\begin{abstract}
Existing hand detection methods usually follow the pipeline of multiple stages with high computation cost, \ie, feature extraction, region proposal, bounding box regression, and additional layers for rotated region detection. In this paper, we propose a new Scale Invariant Fully Convolutional Network (SIFCN) trained in an end-to-end fashion to detect hands efficiently. Specifically, we merge the feature maps from high to low layers in an iterative way, which handles different scales of hands better with less time overhead comparing to concatenating them simply. Moreover, we develop the Complementary Weighted Fusion (CWF) block to make full use of the distinctive features among multiple layers to achieve scale invariance. To deal with rotated hand detection, we present the rotation map to get rid of complex rotation and derotation layers. Besides, we design the multi-scale loss scheme to accelerate the training process significantly by adding supervision to the intermediate layers of the network. Compared with the state-of-the-art methods, our algorithm shows comparable accuracy and runs a $4.23$ times faster speed on the VIVA dataset and achieves better average precision on Oxford hand detection dataset at a speed of $62.5$ fps.
\end{abstract}

\section{Introduction}
Hand detection is applied in many tasks such as virtual reality, human-computer interaction, and driving monitoring, to name a few. However, it is still challenging due to many difficulties such as the low-resolution, clutter background, occlusions, the varying sizes and shapes of hands due to different view angle, and inconsistent appearances due to changing illuminations.

Several methods have been developed for the hand detection in the literature. Traditional methods first employ human-crafted features such as Histograms of Oriented Gradients (HOG)~\cite{2} and skin color~\cite{3}, and then distinguish hand regions by classifiers such as SVM~\cite{12} and Latent SVM~\cite{1}. However, most of these methods are not robust to background clutters, which give rise to high false positive rates.

In recent years, more effective hand detection methods are obtained based on the deep learning based object detection methods, for instance, Region-Based Convolutional Networks (R-CNNs)~\cite{4}, Faster Region-based Convolutional Network (Faster R-CNN)~\cite{5}, and Single Shot MultiBox Detector (SSD)~\cite{6}. Multi-scale features~\cite{8,23} extracted by Convolutional Neural Networks are explored to detect different scales of hands. In the very recent work~\cite{9}, rotation and derotation layers are added to the network to handle rotated hands. One problem with the existing deep learning based hand detection methods is that merging all the multi-scale features rudely and the complex network structure to handle rotation usually lead to high computational cost, which limits the practicality of these methods in applications that require fast processing time.

\begin{figure}[t]
	\centering
	\includegraphics[width=3.2in]{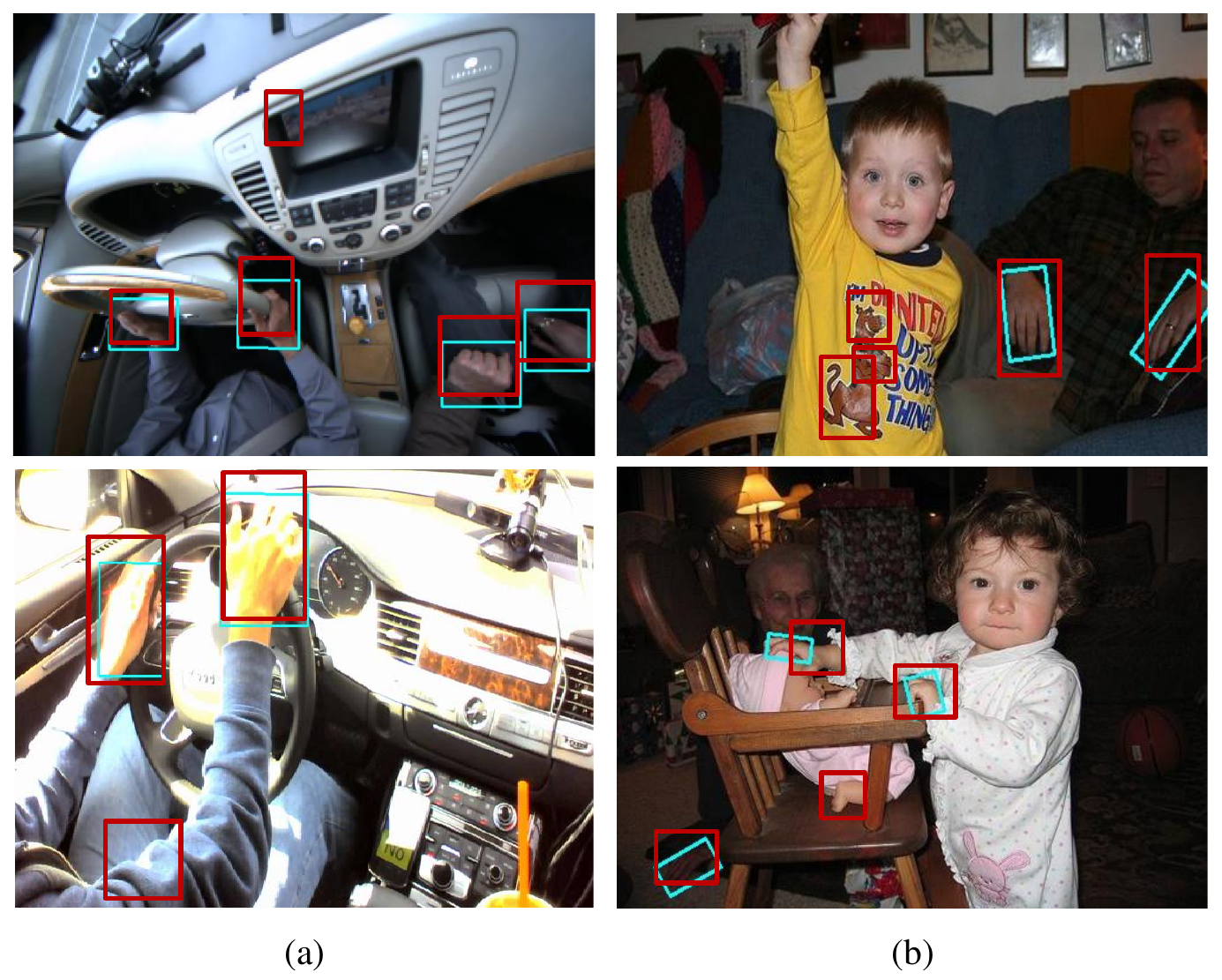}
	\caption{Performance comparison between our SIFCN with ResNet50 backbone (cyan) and Multi-scale fast RCNN~\cite{23} (red). (a) Examples from the VIVA dataset. (b) Examples from the Oxford dataset.}
	\label{compare}
\end{figure}

To address these issues, in this paper, we propose a new efficient hand detection method termed as Scale Invariant Fully Convolutional Network (SIFCN). SIFCN employs existing popular deep learning architecture such as VGG16~\cite{13} or ResNet50~\cite{10} as backbone, and synthesizes multi-scale features to make predictions so as to be scale-invariant for handling hands of different sizes. To reduce the computation cost, we merge feature maps from multiple layers iteratively instead of concatenating them simply. Before that, the $1\times1$ convolution kernel is conducted on feature maps from higher layers to reduce the number of output channels by controlling the number of kernels, as a result of which the computation cost in the next steps is decreased further more. We develop the Complementary Weighted Fusion (CWF) block to make full use of the distinctive features among multiple layers and exploit complementary information. Different from previous methods using additional rotation and derotaion layers~\cite{9}, our model generates the rotation map to represent the rotated hand regions effectively. Moreover, we design the multi-scale loss to accelerate the training process by providing supervision to the intermediate layers of the network. Finally, the Non-Maximum Suppression (NMS) is applied to the bounding boxes detected by our network to yield the detection results. As shown in Figure~\ref{compare}, our SIFCN method detects different scales of hands well. It achieves fewer false positives and generates more accurate hand locations than Multi-scale fast RCNN~\cite{23}. Besides, our model predicts the hand orientation precisely on the Oxford dataset with rotated hand annotations, due to the incorporation of the rotation map. The main contributions of this paper are summarized as follows:
\begin{itemize}
\item We propose a new Scale Invariant Fully Convolutional Network for hand detection, which makes full use of the distinctive features of multiple scales in an iterative way with the CWF block. 
\item We design the multi-scale loss scheme to provide supervision to the intermediate layers of the network, leading to faster convergence of the network.
\item Experiments on VIVA and Oxford datasets show that our method achieves competitive performance with the state-of-the-art hand detection methods but with much improved running time efficiency.
\end{itemize}

\section{Related Work}
\subsubsection{Traditional Methods.} Traditional hand detection methods usually consist of human-crafted feature extraction and classifier training. \cite{32} describes a skin detection based method, which uses contour comparison to find hands from skin areas. However, it is difficult to distinguish between hands and faces well since faces and fists share similar contour shapes. \cite{33} proposes a feature fusion strategy for hand detection in clutter background, but it does not perform well under low resolution and occlusions. \cite{2} uses the HOG features to train a SVM classifier, and extend it with a Dynamic Bayesian Network for better performance. \cite{12} combines a hand shape detector, a context-based detector and a skin-based detector to generate region proposals. Then each proposal is scored to obtain the final results using the SVM classifier. Due to the limitation of hand-crafted features, the performance of these traditional methods is sub-par for practical applications.

\subsubsection{Deep learning based Methods.} Motivated by the good performance of convolutional neural networks (CNNs) in computer vision, many recent hand detection methods are proposed based on CNN models. \cite{7} develops a method combining a candidate region generator and CNNs for hand detections in complex egocentric interactions. Context~\cite{24} is also explored to design the hand detector, which provides extended information of the prevalent hand shapes and locations. However, the additional context cues lead to complicated preprocessing and post-processing. In \cite{8}, the Fully Convolutional Network~\cite{15} is used to generate hand region proposals and then the convolution features are sent to the detection network. In terms of merging multi-scale features into a large feature map, the convolution operations are time-consuming in the later steps. Similarly, \cite{23} concatenates the multi-scale feature maps from the last three pooling layers into a large feature map. Although different receptive fields are taken into account, simple concatenation of feature maps results in high computation overhead.

On the other hand, hands are typically in a rotated pose, and rarely being precisely horizontal or vertical in real scenes. To predict more accurate locations and poses of hands, \cite{9} design a shared network for learning features, a rotation network to predict the rotation angle of region proposals, a derotation layer to obtain axis-aligned rotating feature maps and a detection network for the last classification task. However, the method is of great complexity to handle the rotated distances, even when carefully designed.

Different from the aforementioned deep-learning based methods, our model fuses multi-scale features iteratively and handles hand rotation with the rotation map instead of complex rotation and derotation layers, resulting in comparable accuracy and better efficiency. To be more specific, since the high-level feature maps reflect the global features while the low-level feature maps contains more local information, the feature maps from different scales are weighted before merged, so that the features from multiple scales can complement each other in subsequent process.

\begin{figure*}[t]
	\centering
	\includegraphics[width=\textwidth]{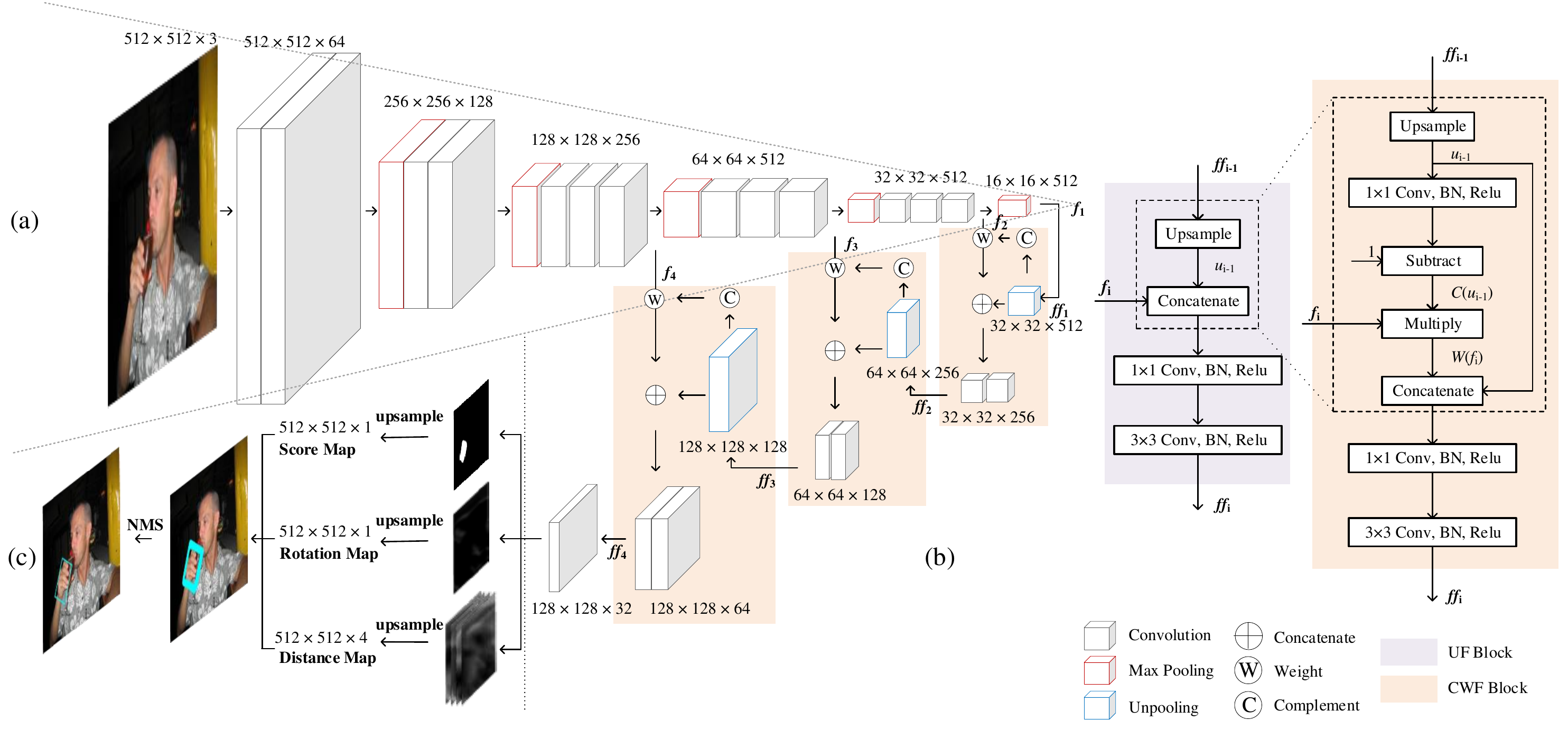}
	\caption{The SIFCN architecture with VGG16 backbone. The network consists of three modules: (a) feature extraction layers, (b) feature fusion layers, and (c) output layers.}
	\label{arch}
\end{figure*}

\section{Scale Invariant Fully Convolutional Network}
As shown in Figure~\ref{arch}, our network is composed of feature extraction layers, feature fusion layers and output layers. In the following, we first describe these modules. Then, we introduce the rotation map to detect rotated hands effectively. Finally, the multi-scale loss function is formulated.

\subsection{Network Architecture}
\subsubsection{Feature Extraction Layers.}
For feature extraction, we employ VGG16~\cite{13} or ResNet50~\cite{10} as the backbone, as shown in Figure~\ref{arch}(a). As multiple scales of hand regions are taken into consideration in the feature fusion layers, we select the feature maps from \textit{pooling-2} to \textit{pooling-5} for VGG16, and \textit{conv2\_1}, \textit{conv3\_1}, \textit{conv4\_1} and \textit{conv5\_1} for ResNet50. The feature maps extracted from VGG16 or ResNet50 are $(\frac{1}{4})^2, (\frac{1}{8})^2, (\frac{1}{16})^2, (\frac{1}{32})^2$ the size of input images, corresponding to the above four scales, respectively. The backbone network is pre-trained based on the ImageNet dataset~\cite{20}.

\subsubsection{Feature Fusion Layers.}
To detect different scales of hands well, it is wise to take full advantage of multi-scale features. However, the computation cost of fusing all feature maps simultaneously can be prohibitive. Therefore, we design the Unweighted Fusion (UF) block to reduce computation overhead, as shown in Figure~\ref{arch}(b), which works as follows:
\begin{itemize}
	\item The last higher level feature maps are up-sampled to fit the size of the current lower level feature maps in the unpooling layer. Then the feature maps from the two levels are concatenated on the channel dimension.
	\item Two convolution operations are performed on the concatenated feature maps. Firstly, the $1\times 1$ convolution is used to reduce the output channels. Then, the $3\times3$ convolution is conducted to combine the feature maps of different scales.
	\item The merged feature maps are regarded as the base feature maps in the next UF block.
\end{itemize}

The UF block is computation-efficient since it merges multi-scale feature maps iteratively. However, it treats the feature maps from different scales equally, \ie, concatenates the current level feature maps with the up-sampled feature maps from the higher layer directly and then conducts convolutions on the concatenated feature maps. Thus the redundant information in the combined features may underestimate the distinctive features, resulting in inferior performance.

To make full use of the distinctive multi-scale features, we introduce Complementary Weighted Fusion (CWF) block as an improved version of UF block. As illustrated in Figure~\ref{arch}(b), the CWF block first weights the current feature maps $f_{s}$ by the up-sampled higher level feature maps $u_{s-1}$ by
\begin{equation}
\label{weighted}
\centering
\left\{
\begin{aligned}
W(f_{s}) &= f_{s} \ast C(u_{s-1}), \\
C(u_{s-1}) &= 1 - \text{Conv}_{1 \times 1}(u_{s-1}).
\end{aligned}
\right.
\end{equation}
$s$ is the current scale, and $W(f_{s})$ is the weighted feature maps. $C(u_{s-1})$ denotes the complementary feature maps of $u_{s-1}$ using the subtraction. $\text{Conv}_{1 \times 1}$ represents the $1 \times 1$ convolution. $\ast$ denotes element-wise multiplication. Weighting $f_{s}$ with $C(u_{s-1})$ can highlight the fine-grained distinctive information contained in $f_{s}$ that $u_{s-1}$ may not have. Then the CWF block concatenates $u_{s-1}$ with $W(f_{s})$ instead of $f_{s}$, so that the feature maps from different scales can fully complement each other. Finally the same two convolutions as used in the UF block are conducted on the concatenated feature maps.

After the final block (\ie, UF block or CWF block), the feature maps go through the $3\times 3$ convolution layer and then are fed to the output layers.

\subsubsection{Output Layers.}
Given the image input, the score map, rotation map and distance map will be generated as the output, as illustrated in Figure~\ref{arch}(c). The width and height of the three kinds of maps are the same as the input image. Similar to the confidence map in Fully Convolutional Networks (FCN)~\cite{15}, each pixel of the score map is a scalar between $0$ and $1$ representing the confidence belonging to a hand region. The rotation map only has $1$ channel recording the rotation angle of the box and the pixel value is in $(-\pi/2, \pi/2)$. Inspired from the work in~\cite{DBLP:conf/cvpr/ZhouYWWZHL17}, the distance map stores the geometry information of hand bounding boxes by $4$ channels, which record the distances to four boundaries of the detected hand bounding box respectively. The bounding boxes of hands are restored with the three kinds of maps and will be purified with the NMS to yield final results.

\subsection{Handling Rotated Hand Detection}
Before performing NMS to remove redundant detection boxes, we restore rotated rectangles from distance maps and rotation maps by estimating the coordinates of the four vertices of the corresponding bounding box for pixels, the scores of which exceed a certain threshold in the score map, called as the score map threshold.
\begin{figure}[htp]
	\centering
	\includegraphics[width=3.2in]{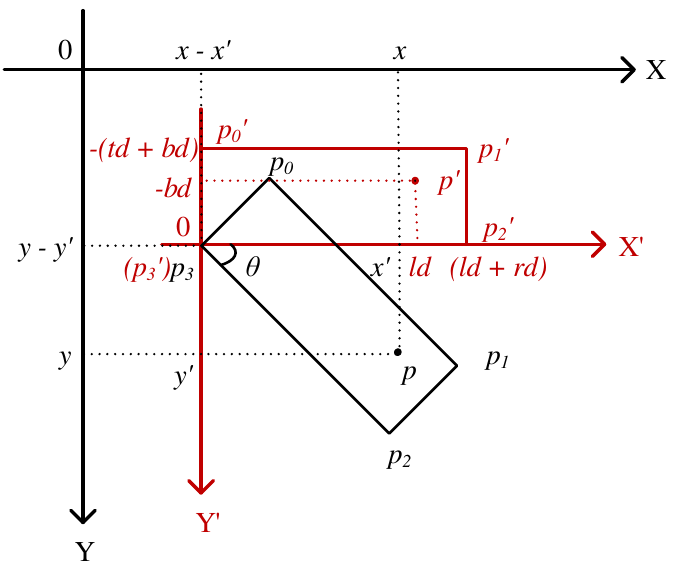}
	\caption{Restore the corresponding rectangle for $p$ from the rotation map and the distance map. The image coordinate system is drawn in black while the auxiliary coordinate system is red.}
	\label{restore}
\end{figure}

To better understand this process, we illustrate with an example of clockwise rotation in Figure~\ref{restore}. Based on the distance map we can obtain the distances $td$, $rd$, $bd$, $ld$ from $p$ to the four boundaries (top, right, bottom, left) of the rectangle $R$. In order to calculate the coordinates of $ p_{0}, p_{1}, p_{2}, p_{3}$ in image coordinate system, an auxiliary coordinate system is introduced with $p_{3}$ as the origin. The directions of X-axis and Y-axis are the same as the image coordinate system. We rotate $R$ to the horizontal around $p_{3}$. $p'$ is the corresponding position of $p$ in the rotated rectangle $R'$. For two-dimensional rotation, the rotation matrix is
\begin{equation}
	\label{r_matrix}
	M\left(\theta \right)=
	\left(
	\begin{array}{cc}
		\cos \theta & -\sin \theta \\
		\sin \theta & \cos \theta
	\end{array}
	\right),
\end{equation}
where $\theta$ is the rotation angle with counter-clockwise as the positive direction, which can be restored from the rotation map. Let $(x', y')$ be the coordinates of $p$ in the auxiliary coordinate system. Then we can calculate the rotation of $p$ as
\begin{equation}
	\label{r_p}
	M\left(\theta \right)
	\left(
	\begin{array}{c}
		x' \\
		y'
	\end{array}
	\right)
	=
	\left(
	\begin{array}{c}
		ld \\
		-bd
	\end{array}
	\right).
\end{equation}
Similarly, for $p_{0}, p_{1}, p_{2}$, we have
\begin{equation}
	\label{r_pi}
	\begin{split}
		M\left(\theta \right)
		\left(
		\begin{array}{c}
			x_{0}' \\
			y_{0}'
		\end{array}
		\right)
		&=
		\left(
		\begin{array}{c}
			0 \\
			-(td+bd)
		\end{array}
		\right), \\
		M\left(\theta \right)
		\left(
		\begin{array}{c}
			x_{1}' \\
			y_{1}'
		\end{array}
		\right)
		&=
		\left(
		\begin{array}{c}
			ld+rd \\
			-(td+bd)
		\end{array}
		\right), \\
		M\left(\theta \right)
		\left(
		\begin{array}{c}
			x_{2}' \\
			y_{2}'
		\end{array}
		\right)
		&=
		\left(
		\begin{array}{c}
			ld+rd \\
			0
		\end{array}
		\right),
	\end{split}
\end{equation}
where $(x_{i}', y_{i}'),\ i\in\{0, 1, 2\}$ are the coordinates of $p_{i}$ in the auxiliary coordinate system. Finally, the coordinates $(x_{i}, y_{i}), i\in\{0, 1, 2, 3\}$ of $ p_{i}$ in the image coordinate system are calculated by
\begin{equation}
	\label{cal_ords}
	\begin{split}
		\left(
		\begin{array}{c}
			x_{3}\\
			y_{3}
		\end{array}
		\right)
		&=
		\left(
		\begin{array}{c}
			x\\
			y
		\end{array}
		\right)
		-
		\left(
		\begin{array}{c}
			x'\\
			y'
		\end{array}
		\right), \\
		\left(
		\begin{array}{c}
			x_{i}\\
			y_{i}
		\end{array}
		\right)
		&=
		\left(
		\begin{array}{c}
			x_{i}'\\
			y_{i}'
		\end{array}
		\right)
		+
		\left(
		\begin{array}{c}
			x_{3}\\
			y_{3}
		\end{array}
		\right),\ 
		i \in \{0, 1, 2\}.
	\end{split}
\end{equation}
$(x, y)$ are the coordinates of $p$ in the image coordinate system. According to Equation~\eqref{r_matrix}$\sim$\eqref{cal_ords}, the rectangle corresponding $p$ can be restored from the rotation map and distance map and represented as $R=\{(x_{i},y_{i})|i\in\{0,1,2,3\}\}$.

\begin{figure*}[htp]
	\centering
	\includegraphics[width=\textwidth]{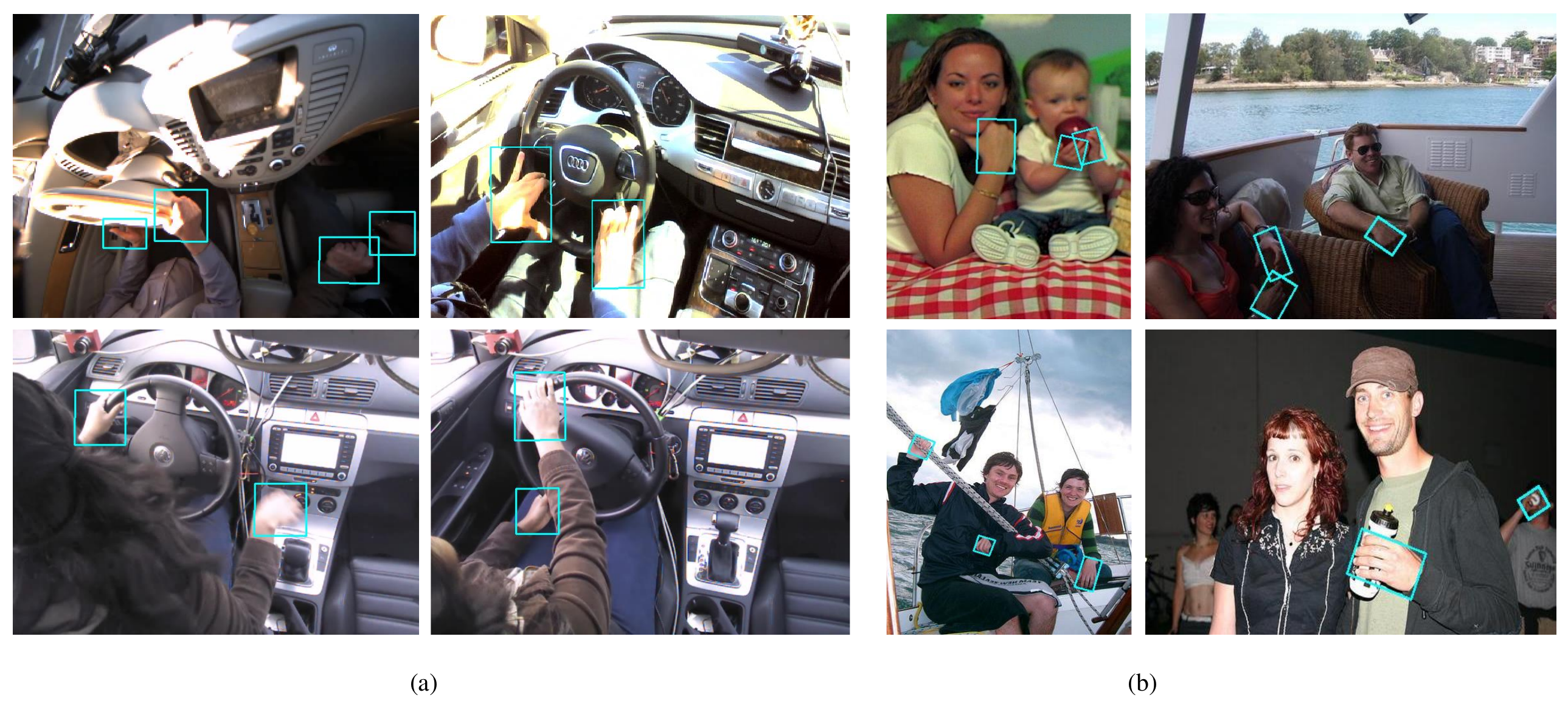}
	\caption{Detection examples of SIFCN with ResNet50 backbone. (a) Examples from the VIVA dataset. (b) Examples from the Oxford dataset.}
	\label{correct_spls}
\end{figure*}

\subsection{Multi-Scale Loss Function}
As discussed above, the output of the network includes three components, namely the score map, the rotation map and the distance map. For better optimization of our model, we add supervision to the intermediate layers in addition to the top output layer. The total loss, named the multi-scale loss, can be calculated as follows:
\begin{equation}
	\label{loss}
	L = \sum\limits_{s \in S}w_{s}\left(\alpha L_{sco}+ \beta L_{rot} + L_{dis}\right),
\end{equation}
where $L_{sco}$, $L_{rot}$ and $L_{dis}$ are losses for the score map, rotation map and distance map, respectively. The scale set $S=\{1,2,3,4\}$ represents the scale index of the extracted feature maps. The parameter $w_{s}$ adjusts the weight of the corresponding scale. The factors $\alpha$ and $\beta$ control the weights of the three loss terms. We explain the three loss functions in detail as follows.

\subsubsection{Loss Function of Score Map}
The dice loss is proved to perform well in segmentation tasks to handle the imbalance problem of positive and negative samples~\cite{16,17,18}. Motivated by this strategy, the loss for the score map can be written as:
\begin{equation}
	\label{s_loss}
	L_{sco} = \dfrac {2 \sum \nolimits_i^N{p_ig_i}}{\sum \nolimits_i^N{p_i^2}+\sum \nolimits_i^N{g_i^2}},
\end{equation}
where the sums run over the all $N$ pixels of the score map. $p_i$ is the value of the pixel $i$ in the score map generated by the prediction network, and $g_i$ is the value of pixel $i$ in the ground truth map.

\subsubsection{Loss Function of Rotation Map}
For the rotation angle, we use cosine function to evaluate the distance between the predicted angle $\tilde{\theta}$ and the ground truth $\theta$:
\begin{equation}\label{r_loss}
	\centering
	L_{rot}= 1-\cos\left(\tilde{\theta}-\theta \right).
\end{equation}

\subsubsection{Loss Function of Distance Map}
Since the loss of distance maps should be scale-invariant, the IoU loss~\cite{27} is adopted to calculate the loss of distance:
\begin{equation}\label{d_loss}
	\centering
	L_{dis}=-\log\dfrac{\tilde{X} \cap X}{\tilde{X} \cup X},
\end{equation}
where $\tilde{X}$ and $X$ are the predicted axis-aligned box and the ground truth bounding box, respectively.

\begin{table*}[htb]
	\caption{Results on VIVA Dataset.}
	\label{VIVAResults}
	\resizebox{\textwidth}{!}{
		\begin{tabular}{l lll l}
			\hline\noalign{\smallskip}
			Methods & Level-1(AP/AR)/\% & Level-2(AP/AR)/\% & Speed/fps & Environment\\
			\noalign{\smallskip}\hline\noalign{\smallskip}
			MS-RFCN~\cite{8} & \textbf{95.1/94.5} & 86.0/\textbf{83.4} & 4.65 & \multirow{2}{*}{6 cores@3.5GHz, 32GB RAM, Titan X GPU}\\
			MS-RFCN~\cite{22} & 94.2/91.1 & \textbf{86.9}/77.3 & 4.65 & \\
			Multi-scale fast RCNN~\cite{23} & 92.8/82.8 & 84.7/66.5 & 3.33 & 6 cores@3.5GHz, 64GB RAM, Titan X GPU\\
			FRCNN~\cite{24} & 90.7/55.9 & 86.5/53.3&-&-\\		
			YOLO~\cite{25} & 76.4/46.0 & 69.5/39.1 & 35.00 & 6 cores@3.5GHz, 16GB RAM, Titan X GPU\\
			ACF\_Depth4~\cite{11} & 70.1/53.8 & 60.1/40.4&-&-\\
			\noalign{\smallskip}\hline\noalign{\smallskip}
			Ours (VGG16+UF) & 88.9/82.8 & 72.6/56.7 & 13.88 &\multirow{6}{*}{4 cores@4.0GHz, 32GB RAM, GeForce GTX 1080}\\
			Ours (VGG16+UF+Multi-Scale Loss) & 92.9/88.3 & 80.9/62.7 & 13.16 &\\
			Ours (VGG16+CWF+Multi-Scale Loss) & 92.3/89.1 & 83.6/68.8 & 13.10 &\\
			Ours (ResNet50+UF) & 93.7/89.9 & 83.6/73.6 & 20.40 &\\
			Ours (ResNet50+UF+Multi-Scale Loss) & 94.0/90.1 & 85.7/74.0 & 20.00 &\\
			Ours (ResNet50+CWF+Multi-Scale Loss) & \textbf{94.6}/\textbf{92.1} & \textbf{86.3}/\textbf{75.8} & 19.68 &\\
			\noalign{\smallskip}\hline
	\end{tabular}}
\end{table*}

\begin{table}[htp]
	\caption{Results on Oxford Dataset.}
	\label{OxfordResults}
	\centering
	\resizebox{.95\columnwidth}{!}{
		\begin{tabular}{ll}
			\hline\noalign{\smallskip}
			Methods & AP/\%\\
			\noalign{\smallskip}\hline\noalign{\smallskip}
			MS-RFCN~\cite{8} & \textbf{75.1}\\
			Multiple proposals~\cite{12} & 48.2\\
			Multi-scale CNN~\cite{23} & 58.4\\
			\noalign{\smallskip}\hline\noalign{\smallskip}
			Ours (VGG16+UF) & 68.7\\
			Ours (VGG16+UF+Multi-Scale Loss) & 77.8\\
			Ours (VGG16+CWF+Multi-Scale Loss) & 78.0\\
			Ours (ResNet50+UF) & 78.2\\
			Ours (ResNet50+UF+Multi-Scale Loss) & 78.6\\
			Ours (ResNet50+CWF+Multi-Scale Loss) & \textbf{80.4}\\
			\noalign{\smallskip}\hline
		\end{tabular}
	}
\end{table}

\section{Experiments}\label{experiment}
We evaluate our algorithm\footnote{The source code of the proposed method is available at \url{http://39.107.81.62/Diana/sifcn}.}, and compare it with existing methods on two benchmark datasets: the VIVA hand detection dataset~\cite{11} and the Oxford hand detection dataset~\cite{12}. We show several qualitative examples in Figure~\ref{correct_spls}. As these results show, the SIFCN with ResNet50 backbone can handle different scales of hands and shapes in various illumination conditions, even the blurred samples.

\subsection{Dataset}
\textbf{VIVA Hand Detection Dataset} is used in the Vision for Intelligent Vehicles and Applications Challenge~\cite{11}. The training set includes $5,500$ annotated images, and the testing set with ground truths that is publicly accessed includes the same number images. The images are extracted from $54$ videos collected in naturalistic driving scenarios. Annotations are given in \textit{.txt} format. The bounding boxes of hand regions are represented as $(x, y, w, h)$, where $x, y$ are the upper-left coordinates of the box and $w$, $h$ are the width and height of the box, respectively. Note that, the annotations are axis-aligned so that the rotation angles are set to $0$ in training and the predictions are axis-aligned bounding boxes in our experiments.

\noindent \textbf{Oxford Hand Detection Dataset} consists of three parts: the training set, the validation set and the testing set, with $1,844$, $406$ and $436$ images separately. Unlike the VIVA dataset, the images in Oxford dataset are collected from various different scenes. Moreover, the ground truth is given by the four vertexes $(x_i, y_i), i \in \{1,2,3,4\}$ of the box in the format of \textit{.mat} and not necessarily to be axis aligned but oriented with respect to the wrist. The rotation angle will be calculated furthermore in our experiments.

\subsection{Experimental Settings}
The experiments are conducted on a single GeForce GTX 1080 GPU and an Intel(R) Core(TM) i7-6700K @ 4.00GHz CPU. For comprehensive evaluation, we try two backbone networks: VGG16~\cite{13} and ResNet50~\cite{10} with the ImageNet~\cite{20} pre-trained models.

Training is implemented with stochastic gradient algorithm using the ADAM scheme. We take the exponential decay learning rate, the initial value of which is $0.0001$ and decays every $10,000$ iterations with rate $0.94$. $w_{s}, s\in\{1,2,3,4\}$ are all set to $1$. The hyper-parameters $\alpha$, $\beta$ are set to $0.01$ and $20$, respectively. Besides, the score map threshold is set to $0.8$ and the NMS is conducted with a threshold $0.2$.

For data augmentation, we randomly mirror and crop the images, as well as do color jittering by distorting the hue, saturation and brightness. Due to the limitation of the GPU capacity, the batch size is set as $12$ and all the images are resized to $512\times 512$ before fed into the network.

\subsection{Evaluations on VIVA Dataset}
Following the Vision for Intelligent Vehicles and Applications Challenge, we evaluate the algorithms on two levels according to the size of the hand instances. Specifically, \textit{Level-1} evaluates the instances with minimum height of $70$ pixels while \textit{Level-2} with $25$ pixels. The Average Precision (AP) and Average Recall (AR) are used to rank compared methods on the VIVA dataset. AP is the area under the precision-recall curve and AR is calculated over $9$ evenly sampled points in log space between $10^{-2}$ and $10^{0}$ false positives per image. As performed in PASCAL VOC~\cite{27}, the hit/miss threshold of the overlap between a pair of predicted and ground truth bounding boxes is set to $0.5$.

As presented in Table~\ref{VIVAResults}, we compare our methods with MS-RFCN~\cite{8,22}, Multi-scale fast RCNN~\cite{23}, FRCNN~\cite{24}, YOLO~\cite{25} and ACF\_Depth4~\cite{11}. Using VGG16 as the backbone network, our model achieves $92.3\%/89.1\%$ (AP/AR) at \textit{Level-1} while $83.6\%/68.8\%$ (AP/AR) at \textit{Level-2}. In terms of ResNet50, we obtain more accurate performance, \ie, $94.6\%/92.1\%$ (AP/AR) at \textit{Level-1} and $86.3\%/75.8\%$ (AP/AR) at \textit{Level-2}. Besides, the running speeds of SIFCN based on VGG16 and ResNet50 are $13.10$ and $19.68$ fps, respectively. 

YOLO~\cite{25} performs hand detection in real time, but its accuracy is unsatisfactory. On the contrary, MS-RFCN~\cite{8} performs against other competitors in accuracy but the detecting speed is very slow, \ie, $4.65$ fps. Therefore, it is of great significance that our model achieves a good trade-off between the accuracy and speed. The model (ResNet50+CWF+Multi-Scale Loss) is comparable to~\cite{8} in accuracy while achieves a $4.23$ times faster running speed as shown in Table~\ref{VIVAResults}.

\subsection{Evaluations on Oxford Dataset}
According to the official evaluation tool\footnote{\url{http://www.robots.ox.ac.uk/~vgg/data/hands/index.html}} in the Oxford dataset, we report the performance on all the ``bigger'' hand instances, those with more than $1,500$ pixels. As shown in Table~\ref{OxfordResults}, similar to the results on VIVA dataset, ResNet50 performs better than VGG16 as a backbone network. Specifically, ResNet50 based SIFCN achieves an improvement of $5.3\%$ in AP score compared with the state-of-the-art MS-RFCN~\cite{8}. VGG16 based SIFCN still outperforms MS-RFCN~\cite{8} by $2.9\%$ in AP score. In addition, it is worth mentioning that the detecting speed on the Oxford dataset is up to $62.5$ fps using ResNet50 while $52.6$ fps using VGG16.

\begin{figure*}[htp]
	\centering
	\includegraphics[width=6.5in]{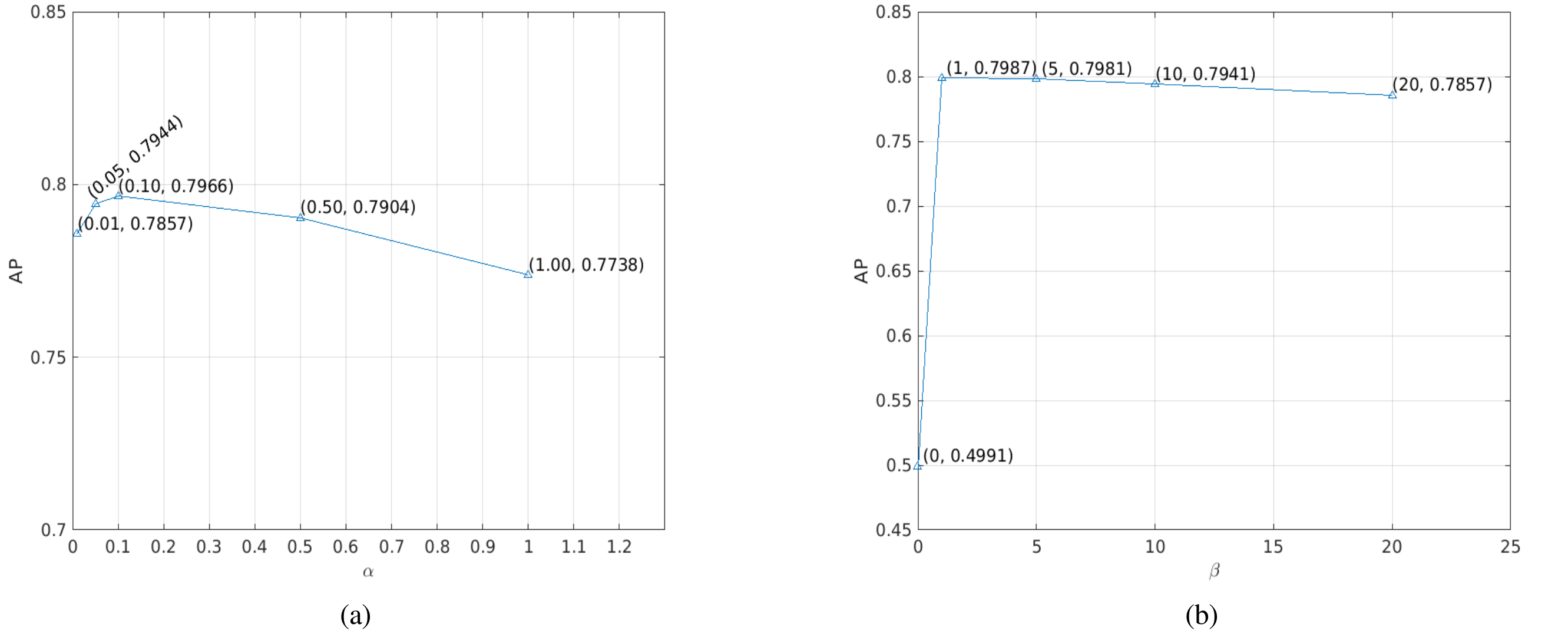}
	\caption{The change of AP with $\alpha$ and $\beta$ on the Oxford dataset. (a) AP score vs. $\alpha$ if $\beta=20$. (b) AP score vs. $\beta$ if $\alpha = 0.01$.} 
	\label{parms}
\end{figure*}
\begin{figure}[htp]
	\centering
	\includegraphics[width=3in]{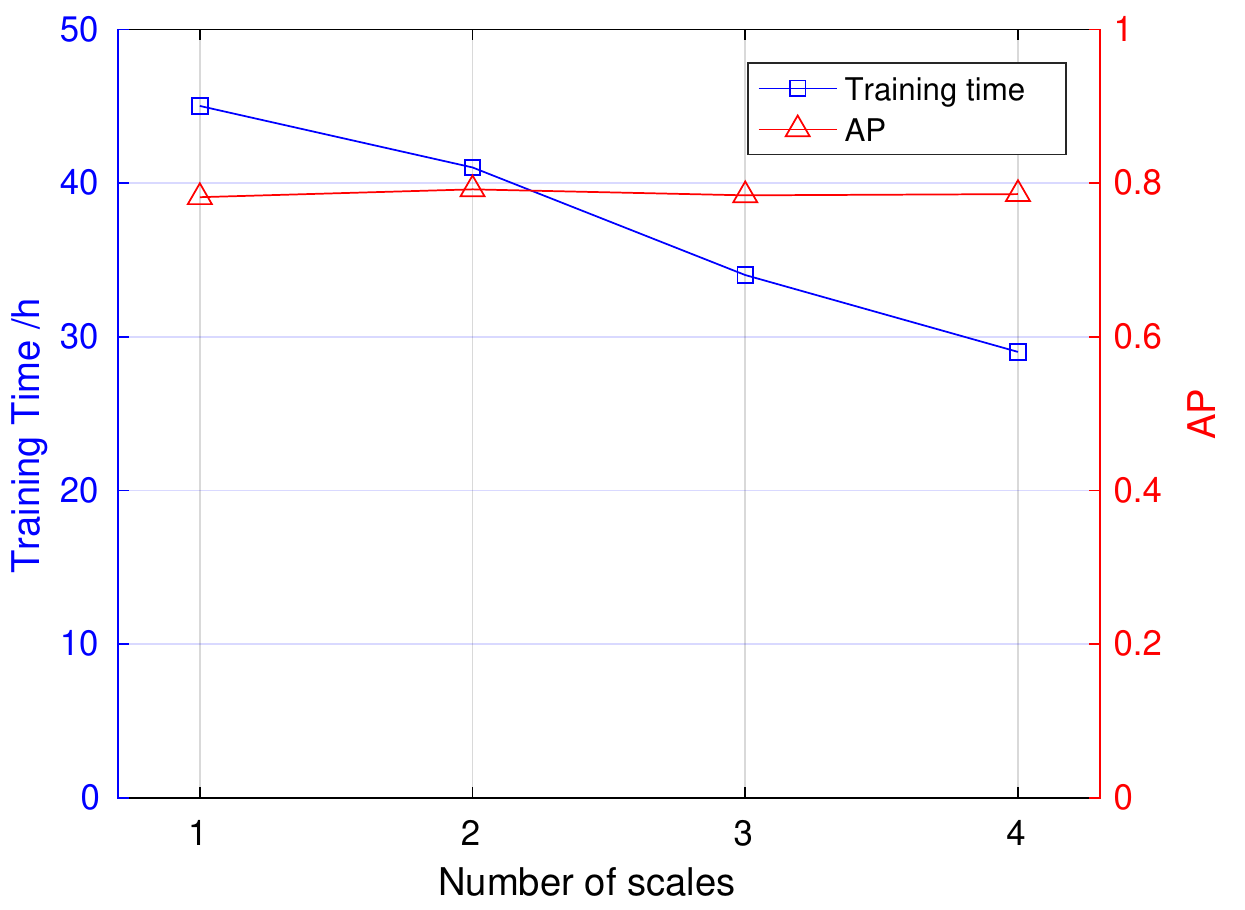}
	\caption{Training time and AP score vs. different numbers of scales for the Oxford dataset.} 
	\label{time}
\end{figure}

\subsection{Ablation Study}
We further perform experiments to study the effect of different aspects of our model on the detection performance. We choose the Oxford dataset to conduct the ablation experiments with the ResNet50 and UF block as the defaults.

{\noindent \textbf{Effectiveness of multi-scale loss.}} In order to investigate the effectiveness of the multi-scale loss, we report the training time and AP scores considering different numbers of scales in Figure~\ref{time}. The number of scales $1, 2, 3, 4$ correspond to $ S=\{4\}, S=\{3,4\}, S=\{2,3,4\}, S=\{1,2,3,4\}$ in Equation~\eqref{loss} respectively. It can be seen that as the number of scales used in loss function increases, the time it takes to train the model to convergence decreases. The convergence of the network is accelerated significantly (more than $10$ hours) using the multi-scale loss. At the same time, there is even a slight increase in AP score. That is, the multi-scale loss accelerates the training process without sacrificing the AP score. This is attributed to the multiple supervision to the intermediate layers of the network.

{\noindent \textbf{Influence of score map.}} We change the value of $\alpha$ in Equation~\eqref{loss} to find appropriate weights of score map in training. The results are reported in Figure~\ref{parms}(a). As $\alpha$ increases from $0.01$ to $1$, the AP increases first and then decreases, and reaches the maximum when $\alpha$ is $0.10$ in our experiments. It can be seen that the AP is not too sensitive to the weight of score map.

{\noindent \textbf{Effectiveness of rotation map.}} As discussed above, $\beta$ in Equation~\eqref{loss} weights the loss of rotation angle in the training process. As shown in Figure~\ref{parms}(b), when the angle loss is considered in the optimization procedure, \ie, $\beta>0$, the AP score is stable and larger than $0.78$ for different values of $\beta$. Otherwise, if $\beta=0$, there is a significant drop in the AP score on Oxford dataset (\ie, $0.4991$). It can be concluded that the rotation map plays a very important role in optimizing the final model.

{\noindent \textbf{Effectiveness of CWF block.}} From Table~\ref{VIVAResults} and \ref{OxfordResults}, we can see that the CWF block outperforms the UF block whether using the VGG16 or ResNet50 as the backbone. Specifically, the CWF block achieves higher AP and AR on VIVA dataset, especially the AR score, which has been greatly improved. It indicates that the model with the CWF block produces less false negatives than the UF block and makes better use of the distinctive features of different scales. For example, the CWF block gains an improvement of $0.2\%$ in AP score with VGG16 and $1.8\%$ with ResNet50 comparing to the UF block on the Oxford dataset.

\section{Conclusion}
We present an efficient Scale Invariant Fully Convolutional Network (SIFCN) for hand detection. The proposed Complementary Weighted Fusion (CWF) block can make full use of the distinctive features of different scales to achieve scale invariance effectively. Specifically, the multi-scale features are merged iteratively rather than concatenated simultaneously to reduce computation overhead. Moreover, the multi-scale loss scheme is employed to accelerate the training procedure significantly. Experimental results on the VIVA and Oxford datasets show comparable performance of our method compared with the state-of-the-art methods with much higher speed. For the future work, we will optimize the code to further improve the run time efficiency of SIFCN so that it can run in real-time with better accuracy.

\bibliography{myreferences}
\bibliographystyle{aaai}

\end{document}